\documentclass[11pt]{article}

\usepackage[preprint]{acl}

\usepackage{times}
\usepackage{latexsym}

\usepackage[T1]{fontenc}

\usepackage[utf8]{inputenc}

\usepackage{microtype}

\usepackage{inconsolata}

\usepackage{graphicx}
\usepackage{array}
\usepackage{makecell}
\usepackage{bm}
\usepackage{multirow}
\usepackage{amsfonts}
\usepackage{xcolor} 
\usepackage{colortbl} 
\usepackage{booktabs}
\usepackage{url}
\usepackage[normalem]{ulem}
\usepackage[most]{tcolorbox}
\usepackage{adjustbox}
\usepackage{tabularx}

\title{PodBench: A Comprehensive Benchmark for Instruction-Aware Audio-Oriented Podcast Script Generation}

\author{
  Chenning Xu, Mao Zheng, Mingyu Zheng, Mingyang Song\thanks{Corresponding author.} \\
  Large Language Model Department, Tencent \\
  \texttt{\{chenningxu,moonzheng,mingyuzheng,nickmysong\}@tencent.com}
}

\begin{document}
\maketitle
\begin{abstract}
Podcast script generation requires LLMs to synthesize structured, context-grounded dialogue from diverse inputs, yet systematic evaluation resources for this task remain limited. To bridge this gap, we introduce PodBench, a benchmark comprising 800 samples with inputs up to 21K tokens and complex multi-speaker instructions. We propose a multifaceted evaluation framework that integrates quantitative constraints with LLM-based quality assessment. Extensive experiments reveal that while proprietary models generally excel, open-source models equipped with explicit reasoning demonstrate superior robustness in handling long contexts and multi-speaker coordination compared to standard baselines. However, our analysis uncovers a persistent divergence where high instruction following does not guarantee high content substance. PodBench offers a reproducible testbed to address these challenges in long-form, audio-centric generation. The code and data are publicly accessible here.\footnote{\url{https://github.com/xucncn/PodBench}}

\end{abstract}

\section{Introduction}
Rapid advancements in large language models (LLMs) have revolutionized generative writing tasks, especially long-form writing~\cite{wu2025writingbench,bai2024longwriter,writing_RL,writing_zero}, supporting more than 40\% of language model interactions~\cite{openai2025usage,anthropic2025economic} and showing strong performance across genres such as brand narrative, creative fiction, and persuasive essays. Among these tasks, podcast script generation, which composes natural podcast dialogue episodes based on user instructions and textual materials to provide constructive discussions and valuable insights, has received increasing attention from both academia and industry due to its broad applications in AI-Podcast products such as Google’s NotebookLM~\cite{notebooklm} and MoonCast~\cite{mooncast}.

However, unlike common generative writing tasks that have been widely studied and evaluated with abundant benchmarks, podcast script generation capabilities of LLMs remain largely under-explored. This is partly due to the lack of a comprehensive and specialized benchmark for evaluating LLMs’ writing performance under real-world AI-Podcast scenarios, which are characterized by complex and diversified user queries and long contextual input materials. Moreover, podcast script writing not only requires general writing qualities such as fluency and coherence, but also involves task-specific constraints, including multi-speaker conversational structure and strict grounding in source materials. Most existing general-purpose writing benchmarks rely on generic, instruction-derived criteria applied uniformly across tasks. In podcast script generation, instruction adherence and listener-oriented script quality reflect related but distinct aspects of performance, which are not always jointly captured by such evaluations.

To bridge this gap, we introduce \textbf{PodBench}, a comprehensive writing benchmark for instruction-aware and context-grounded podcast script generation. Our benchmark comprises 800 test samples and is characterized by two key features. \textbf{(1) Context-grounded}. Similar to real-world AI-Podcast applications, PodBench provides input materials as contextual grounding for script generation, including user-uploaded documents and retrieved web or news articles. We collect diverse materials from publicly available corpora, such as LongWanjuan~\cite{longwanjuan} and PileArxiv~\cite{the_pile}, covering both single-document and multi-document scenarios across 12 common AI-Podcast domains (e.g., social culture, business and finance, education). This enables systematic evaluation of whether LLMs can generate appropriate and hallucination-free podcast scripts across domains. \textbf{(2) Instruction-aware}. Drawing from authentic user instructions in real-world AI-Podcast applications, we identify eight important requirement types for podcast creation, such as podcast language, speaker number, focused content, and script structure. Based on these requirement dimensions and the input materials, we leverage LLMs to synthesize diverse podcast instructions, ranging from simple prompts with limited constraints to complex instructions composed of multiple demands. PodBench not only requires models to produce lengthy podcast scripts, but also includes input prompts reaching 21K tokens, thus challenging LLMs’ capabilities in both long-context understanding and generation. Furthermore, we develop an evaluation framework that distinguishes instruction following from podcast-specific script quality, combining deterministic metrics with LLM-based assessment to better support audio-oriented dialogue generation.

Using PodBench, we evaluate a wide spectrum of LLMs, including mainstream open-source models, closed-source advanced models, and writing-enhanced specialist models. The results reveal clear performance differences across model families: proprietary models consistently achieve the strongest overall results, while recent open-source models show notable improvements in instruction following and structured dialogue generation. At the same time, challenges related to long-context understanding, multi-speaker interaction, and content depth continue to expose meaningful variation in model capabilities.

We conclude our contributions as follows:
 \begin{itemize}
    \item Targeted at the gap of existing LLM writing benchmarks and the rapidly growing applications of AI-Podcast, we introduce a new benchmark PodBench to evaluate LLMs' proficiency in composing podcast scripts based on diverse user queries and input materials.
    \item Considering the limitations of traditional evaluation methods, we establish a multifaceted evaluation framework, integrating deterministic quantitative measures with LLM-based assessment.
    \item On this basis, we conduct a thorough evaluation of various types of LLMs, revealing their pros and cons in the podcast script generation task and providing valuable insights for future AI-podcast applications
\end{itemize}

\section{Related Work}

\subsection{AI-Podcast}
The rapid progress of AI-Generated Content (AIGC) has enabled a wide range of applications, including advanced chatbots~\cite{gpt4_tech_report} and multimodal image generation~\cite{seedream_4}. More recently, AI-Podcast has emerged as a distinct audio-first paradigm for information consumption, offering hands-free and eyes-free access to content in scenarios such as driving or household activities~\cite{notebooklm,seed_ai_podcast_model,ListenHub}.

Following the introduction of Google’s NotebookLM~\cite{notebooklm}, research on AI-Podcast has grown rapidly and can be broadly categorized into two directions. The first explores end-to-end audio LLMs or omni-modal models that directly generate podcast audio via joint text–speech pre-training~\cite{seed_ai_podcast_model,openai2024gpt4ocard}, aiming to tightly integrate language understanding and speech synthesis. The second, more widely adopted approach constructs agentic pipelines by coupling LLM-based script generation with cascaded TTS modules~\cite{podagent,mooncast,WavJourney}. In this setting, LLMs first generate structured podcast scripts, which are then rendered into audio with speech synthesis and sound effects. Representative systems include MoonCast~\cite{mooncast}, which uses Gemini models for document understanding and script generation, and PodAgent~\cite{podagent}, which proposes a Host--Guest--Writer agent framework for dialogue composition.

Due to current limitations of audio LLMs in producing long-form speech directly, most practical AI-Podcast systems rely on such script-centered pipelines, making the quality of podcast script generation a critical factor in overall podcast performance.

\subsection{LLM-based Generative Writing}
Recent advances in LLMs have substantially improved long-form text generation, motivating a series of benchmarks for evaluating generative writing quality~\cite{wu2025writingbench, hellobench, bai2024longwriter, paech2023eqbench}. For example, WritingBench~\cite{wu2025writingbench} evaluates diverse writing tasks across multiple domains using LLM-synthesized queries and LLM-as-a-judge evaluation, a common practice in the absence of ground-truth references. Alongside benchmarking efforts, prior work has explored improving writing capability through domain-specific pre-training~\cite{weaver}, high-quality supervised fine-tuning and distillation~\cite{wu2025writingbench}, instruction back-translation~\cite{pham-etal-2024-suri}, agent-based generation pipelines~\cite{bai2024longwriter, tu2025longwriterv, REER, self_lengthen}, and writing-oriented reinforcement learning~\cite{writing_RL, writing_zero, long_writer_zero}.

Despite this progress, podcast script generation remains relatively under-explored, largely due to the absence of specialized benchmarks. PodBench aims to bridge this gap by focusing on instruction-aware, context-grounded podcast script generation. The closest related work is PodEval~\cite{podeval}, which targets podcast audio generation with a multimodal evaluation framework and a limited set of script-level prompts. In contrast, PodBench emphasizes more diverse instructions and longer contextual inputs, making it complementary to PodEval.

\section{PodBench}

\begin{figure*}[t]
  \centering
  \includegraphics[width=\textwidth]{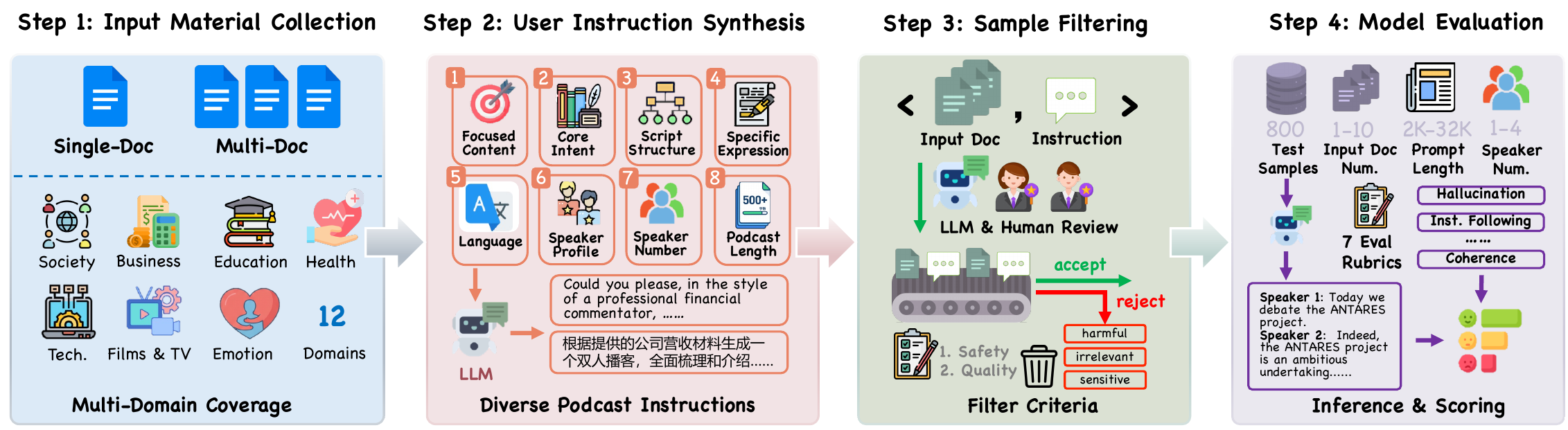}
  \caption{The construction pipeline of PodBench.}
  \label{benchmark_construction_pipeline}
\end{figure*}

\subsection{Task Definition}
Given a user instruction $Inst$ including specific requirements about expected podcast and a set of input materials $Doc$ consisting of several documents, the podcast script generation task requires the model $f(\cdot)$ to compose podcast dialogue episodes $R$ from specific speakers, i.e., $R=f(Inst,Doc)$. Previous studies mainly focus on how to generate two-speaker podcast solely based on user queries or input documents~\cite{mooncast,podagent}. By constrast, we retain the task setting of real-world AI-podcast application scenarios where users can provide specific requirements and corresponding input materials.

\subsection{Benchmark Construction}
\paragraph{Input Material Collection.} A primary function of AI-Podcast is to assist users in comprehending input materials of diverse domains, such as economics, education, society. To align our study with realistic application settings, we collect documents covering a wide range of topics from public corpora. Chinese document sources include LongWanjuan~\cite{longwanjuan}, OpenNewsArchive~\cite{OpenDataLab} and WanJuan~\cite{WanJuan_1} and English sources include WanJuan-CC~\cite{WanJuan_CC} and Pile Arxiv~\cite{the_pile}. To synthesize multi-document materials, we first employ Gemini-2.5-pro to generate a topic query for each document, and perform document clustering based on the query representation embedding from Qwen3-Embedding-4B, yielding clusters of relevant documents centered on specific topics. 

\paragraph{User Instruction Synthesis.} To capture the diversity of user requirements, we extract 8 key requirement dimensions based on realistic AI-Podcast scenarios.

\noindent \textbf{1. Focused Content.} The target topics or content within input materials that users want the model to emphasize in its discussion. For instance, users may request a comprehensive coverage of the entire document, or alternatively, to elaborate specifically on certain sub-sections within the document.

\noindent \textbf{2. Core Intent.} The primary purpose behind users' engagement with AI podcasts.  For example, users may expect the model to conduct an in-depth discussion on a specific topic, seek a concise summary of the provided documents, or require the model to offer critical perspectives on a subject.

\noindent \textbf{3. Script Structure.} The required discussion logic of the podcast script. Examples include the "general-specific-general" structure, where key questions are first posed, followed by point-by-point discussions, and concluded with a summary, as well as chronological structures that progressively unfold the discussion along the timeline.

\noindent \textbf{4. Specific Expression.} The particular linguistic expressions that users mandate for use within the podcast, e.g., employing domain-specific AI terminology during discussions, utilizing predetermined prologues and concluding statements, and designating specific names for the resulting podcast.

\noindent \textbf{5. Podcast Language.} The required language used in the podcast. For simplicity, we consider Chinese and English podcast.

\noindent \textbf{6. Speaker Profile.} The persona settings of different speakers employed in AI podcast generation, which encompasses the speaker's personality characteristics, conversational style, domain expertise, and assigned roles and responsibilities.

\noindent \textbf{7. Speaker Number.} The number of podcast speakers. Unlike previous work that mainly focus on two-speaker podcast, we do not fix the speaker number and allow users to create customized podcast with different speakers, e.g., solo-podcast with one speaker or roundtable discussion podcast with multiple speakers.

\noindent \textbf{8. Podcast Length.} Generating text of a specified length constitutes a critical capability in LLMs' writing tasks. Accordingly, we allow users to specify the desired podcast length. For instance, a podcast involving in-depth exploration of a particular topic may extend to 8,000 words, whereas a podcast providing a brief overview of a document may comprise only around 1,000 words.

Based on the above requirement dimensions, we prompt Gemini-2.5-pro with in-context examples to synthesize diverse user instructions that encompass various numbers of requirement dimensions and different difficulties.

\paragraph{Sample Filtering.} Following the initial collection, we obtain 2.6K initial instances, which may contain inappropriate documents and erroneous instructions. To ensure data quality, we employ 5 human experts to review sample quality and exclude low-quality instances with the aid of LLMs. Concretely, we first use Claude-4.5-opus to review the document content and identify potentially harmful content (e.g., violence, terrorism, or pornography). The model then evaluates whether each instruction constitutes a suitable user input based on criteria including the existence of sensitive content and instruction-document relevance. Subsequently, human experts perform a second verification of document-instruction pairs, eliminating low-quality instances and revising instructions with minor mistakes. Ultimately, This process results in a final set of 400 Chinese and 400 English samples that align with the input materials and reflect practical AI-Podcast scenarios.

\begin{table*}[t]\footnotesize
\centering
\renewcommand{\arraystretch}{1.2}
\setlength\tabcolsep{11pt}
\scalebox{0.88}{
\begin{tabular}{l|c|c|c|c|c|c} \hline
\multirow{2}{*}{\begin{tabular}[c]{@{}c@{}}\textbf{\textbf{}}\\\textbf{\textbf{Method}}\end{tabular}} & \multirow{2}{*}{\begin{tabular}[c]{@{}c@{}}\textbf{ Instruction }\\\textbf{Following}\\\textbf{(0-100)}\\\end{tabular}} & \multicolumn{4}{c|}{\textbf{Content Quality }} & \multirow{2}{*}{\begin{tabular}[c]{@{}c@{}}\textbf{}\\\textbf{Ave. }\end{tabular}} \\ \cline{3-6}
 &  & \begin{tabular}[c]{@{}c@{}}\textbf{Depth \& Value}\\\textbf{(0-45)}\end{tabular} & \begin{tabular}[c]{@{}c@{}}\textbf{Structure}\\\textbf{(0-30)}\end{tabular} & \begin{tabular}[c]{@{}c@{}}\textbf{Language}\\\textbf{(0-25)}\end{tabular} & \begin{tabular}[c]{@{}c@{}}\textbf{Overall}\\\textbf{(0-100)}\end{tabular} &  \\ \hline
\multicolumn{7}{l}{{\cellcolor[rgb]{0.933,0.933,0.933}}\textit{Proprietary-LLMs}} \\
Claude-4-5-sonnet & \textbf{96.58} & 25.39 & 19.44 & 18.42 & 63.25 & 79.91 \\
GPT-5.1 & 95.52 & \textbf{28.61} & \textbf{21.45} & \textbf{19.58} & \textbf{69.64} & \textbf{82.58} \\
GPT-4o & 89.43 & 22.10 & 18.02 & 16.74 & 56.86 & 73.15 \\
Gemini-3-pro-preview & 96.09 & 25.49 & 19.93 & 18.43 & 63.85 & 79.97 \\
\multicolumn{7}{l}{{\cellcolor[rgb]{0.933,0.933,0.933}}\textit{Open-source LLMs (instruct mode)}} \\
InternLM3-8B-instruct & 73.95 & 18.19 & 14.72 & 12.98 & 45.89 & 59.92 \\
Llama-3.1-8B-Instruct & 49.73 & 13.89 & 11.81 & 11.81 & 37.51 & 43.62 \\
Llama-3.1-70B-Instruct & 58.65 & 15.27 & 12.99 & 12.65 & 40.91 & 49.78 \\
Llama-4-Scout-17B-16E-Instruct & 70.14 & 16.62 & 14.16 & 13.73 & 44.51 & 57.33 \\
Qwen2.5-7B-Instruct & 69.13 & 17.00 & 14.06 & 12.90 & 43.96 & 56.55 \\
Qwen2.5-32B-Instruct & 77.84 & 17.42 & 14.92 & 14.51 & 46.85 & 62.35 \\
Qwen2.5-72B-Instruct & 83.31 & 18.93 & 15.89 & 14.24 & 49.06 & 66.19 \\
Qwen3-1.7B & 56.52 & 15.04 & 12.72 & 12.85 & 40.61 & 48.57 \\
Qwen3-4B & 76.73 & 17.80 & 14.72 & 15.17 & 47.69 & 62.21 \\
Qwen3-8B & 80.79 & 18.81 & 15.41 & 15.70 & 49.93 & 65.36 \\
Qwen3-14B & 87.07 & 19.89 & 16.36 & 15.97 & 52.22 & 69.65 \\
Qwen3-32B & 90.69 & 21.67 & 17.51 & 17.21 & 56.39 & 73.54 \\
Qwen3-30B-A3B & 86.50 & 19.40 & 16.09 & 16.79 & 52.28 & 69.39 \\
Qwen3-235B-A22B & \textbf{93.64} & 22.42 & 18.20 & \textbf{17.77} & 58.39 & 76.02 \\
DeepSeek-V3-0324 & 92.71 & \textbf{23.15} & \textbf{18.70} & 17.76 & \textbf{59.61} & \textbf{76.16} \\
\multicolumn{7}{l}{{\cellcolor[rgb]{0.933,0.933,0.933}}\textit{Open-source LLMs (thinking mode)}} \\
Qwen3-1.7B & 62.65 & 16.40 & 13.29 & 12.60 & 42.29 & 52.47 \\
Qwen3-4B & 83.04 & 19.68 & 15.91 & 15.57 & 51.15 & 67.10 \\
Qwen3-8B & 88.45 & 20.94 & 16.68 & 16.26 & 53.89 & 71.17 \\
Qwen3-14B & 91.79 & 22.37 & 17.80 & 16.70 & 56.86 & 74.33 \\
Qwen3-32B & 93.78 & 23.82 & 18.70 & 17.40 & 59.92 & 76.85 \\
Qwen3-30B-A3B & 89.88 & 21.27 & 16.83 & 16.68 & 54.78 & 72.33 \\
Qwen3-235B-A22B & \textbf{94.32} & 23.98 & 18.83 & 17.62 & 60.42 & 77.37 \\
DeepSeek-R1-Distill-Llama-8B & 66.94 & 16.53 & 14.53 & 14.16 & 45.22 & 56.08 \\
DeepSeek-R1-Distill-Qwen-7B & 42.59 & 13.58 & 11.19 & 11.43 & 36.21 & 39.40 \\
DeepSeek-R1-Distill-Qwen-32B & 80.12 & 17.98 & 15.29 & 14.78 & 48.05 & 64.09 \\
DeepSeek-R1-0528-Qwen3-8B & 85.67 & 19.85 & 16.24 & 15.20 & 51.30 & 68.49 \\
DeepSeek-R1-0528 & 94.27 & \textbf{24.49} & \textbf{19.13} & \textbf{17.53} & \textbf{61.15} & \textbf{77.71} \\
\multicolumn{7}{l}{{\cellcolor[rgb]{0.933,0.933,0.933}}\textit{Writing-enhanced LLMs}} \\
LongWriter-Llama3.1-8B & 59.04 & 16.11 & 13.31 & 11.27 & 40.69 & 49.86 \\
LongWriter-GLM4-9B & 66.00 & 16.60 & 14.02 & 13.42 & 44.04 & 55.02 \\
LongWriter-zero-32B & \textbf{84.91} & \textbf{23.35} & \textbf{18.36} & \textbf{17.57} & \textbf{59.29} & \textbf{72.10} \\ \hline
\end{tabular}
}
\caption{PodBench breakdown evaluation results of 4 types of LLMs. For all metrics, higher scores indicate better performance. `Ave.' denotes the average score of `Instruction Following' and `Overall Content Quality'.} 
\label{main_results_table}
\end{table*}

\subsection{Benchmark Analysis}
\label{sec:benchmark_analysis}
The resulting dataset, named \textbf{PodBench}, comprises 800 test samples. In comparison to existing writing benchmarks summarized in Table \ref{benchmark_comparison}, PodBench is the first comprehensive podcast script generation benchmark which is featured with the following advantages. (1) Broad domain coverage. It contains input materials collected from 12 domains, spanning single- to multi-document scenarios. (2) Diverse podcast instructions. Drawing from practical AI-Podcast applications, we identify 8 important requirement dimensions and synthesize various instructions to simulate real user demand, covering basic writing requirements like length and format, as well as podcast-oriented requirements like speaker profile. (3) Long context grounding. Models are required to understand long input documents with the maximum length up to 21K, and to generate lengthy podcast scripts where key information must be grounded in input materials to avoid hallucination. The key statistics of PodBench are presented in Table~\ref{podbench_statistics_table}, and the domain distribution is illustrated in Figure~\ref{domain_distribution}. More information about PodBench including benchmark examples is given in Appendix~\ref{sec::more_benchmark_info}.

\begin{table*}[t]\footnotesize
    \centering
    \scalebox{0.82}{
    \renewcommand{\arraystretch}{1.2}
\begin{tabular}{l|ccc|cccc|cc|c|c} 
\hline
\multirow{2}{*}{\textbf{Benchmark}} & \multirow{2}{*}{\textbf{Num.}} & \multirow{2}{*}{\textbf{Domains}} & \multirow{2}{*}{\textbf{Lang.}} & \multicolumn{4}{c|}{\textbf{Requirement Dimensions}} & \multicolumn{2}{c|}{\textbf{Input Token}} & \multirow{2}{*}{\begin{tabular}[c]{@{}c@{}}\textbf{Free }\\\textbf{Instruction}\end{tabular}} & \multirow{2}{*}{\begin{tabular}[c]{@{}c@{}}\textbf{Diverse}\\\textbf{Material}\end{tabular}} \\ 
\cline{5-10}
 &  &  &  & \textbf{Style} & \textbf{Format} & \textbf{Length} & \textbf{Speaker} & \textbf{Avg} & \textbf{Max} &  &  \\ 
\hline
EQ-Bench & 241 & 1 & En & \texttimes & \texttimes & \texttimes & \texttimes & 130 & 213 & \texttimes & - \\
LongBench-Write & 120 & 7 & En, Zh & \texttimes & \texttimes & \checkmark & \texttimes & 87 & 684 & \checkmark & - \\
HelloBench & 647 & 5 & En & \texttimes & \texttimes & \checkmark & \texttimes & 1,210 & 7,766 & \texttimes & \texttimes \\
WritingBench & 1,000 & 6 & En, Zh & \checkmark & \checkmark & \checkmark & \texttimes & 1,699 & 19,361 & \checkmark & \checkmark \\
PodEval & 51 & 17 & En & \texttimes & \texttimes & \texttimes & \texttimes & 10 & 18 & \texttimes & - \\
\textbf{PodBench} & 800 & 12 & En, Zh & \checkmark & \checkmark & \checkmark & \checkmark & 4,922 & 21,649 & \checkmark & \checkmark \\
\hline
\end{tabular}
}
    \caption{Comparison of PodBench and existing writing benchmarks. `Num.' and `Lang.' represents sample number and sample language. `Speaker' indicates whether the test samples require models to compose podcast scripts based on customized speaker profiles.}
    \label{benchmark_comparison}
\end{table*}

\begin{table}[t]\footnotesize
    \centering
    \renewcommand{\arraystretch}{1.2}
    \scalebox{0.9}{
    \begin{tabular}{c|c|c|c|c} \hline
\textbf{Characteristic} & \begin{tabular}[c]{@{}c@{}}\textbf{Sample }\\\textbf{Num.}\end{tabular} & \textbf{Mean} & \textbf{Min} & \textbf{Max} \\ \hline
Doc Num. & 800 & 2.7 & 1 & 10 \\
Inst. Length & 800 & 40.7 & 4 & 158 \\
Speaker Num. & 800 & 1.9 & 1 & 4 \\ \hline
\multicolumn{5}{c}{\textbf{Input Prompt Length}} \\ \hline
Overall & 800 & 4922.3 & 1331 & 21649 \\
0-2K & 139 & 1721.6 & 1331 & 2047 \\
2K-4K & 360 & 2957.1 & 2052 & 4094 \\
4K-8K & 153 & 5443.4 & 4100 & 8147 \\
8K-16K & 134 & 11571.6 & 8199 & 16066 \\
16-21K & 13 & 19115.6 & 16570 & 21649 \\ \hline
\end{tabular}
    }
    \caption{Basic statistics of PodBench. `Doc Num.' and `Inst. Length' denotes the input document number and instruction length. The tokenizer of Qwen3-8B is adopted to calculate the token length of input prompt and synthesized instructions.}
    \label{podbench_statistics_table}
\end{table}

\begin{figure}[t]
  \centering
  \includegraphics[width=0.65\linewidth]{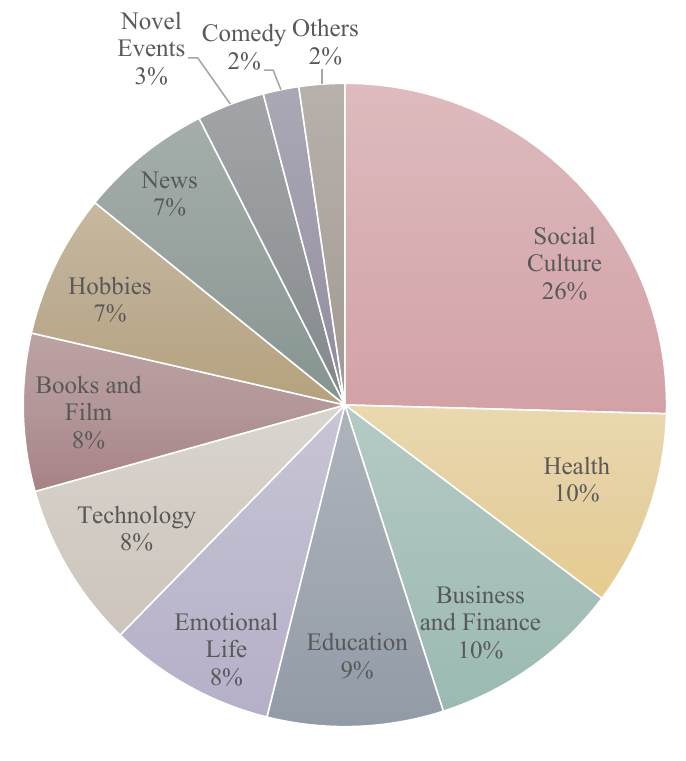}
  \caption{Domain categories of input documents.}
  \label{domain_distribution}
\end{figure}

\subsection{Evaluation Framework}
\label{sec:evaluation_framework}

\begin{figure}[t]
  \centering
  \includegraphics[width=\columnwidth]{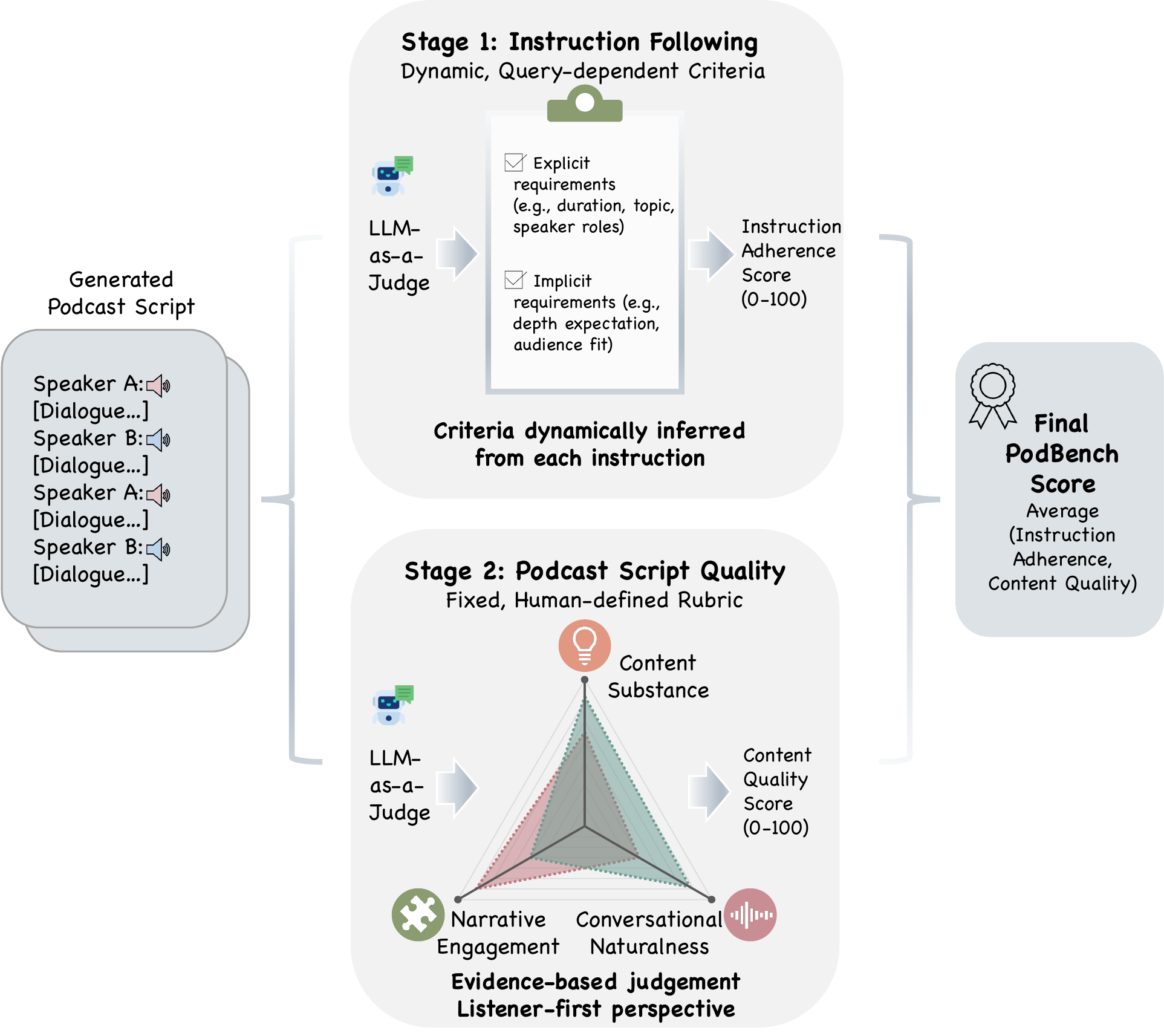}
  \caption{Overview of the evaluation framework.}
  \label{fig:evaluation_framework}
  \vspace{-0.3cm}
\end{figure}

General-purpose writing benchmarks~\cite{wu2025writingbench} typically rely on generic, query-derived checklists for meta-evaluation. While adaptable, this paradigm tends to treat criteria uniformly, potentially overlooking the domain-specific preference weights and domain knowledge inherent to distinct writing scenarios. To address this, we propose a two-stage framework that integrates dynamic constraint checking with a fine-grained, domain-specific rubric. As illustrated in Figure~\ref{fig:evaluation_framework}, this design decouples the assessment of instruction following from domain-grounded, preference-aligned script quality, aiming to better approximate human judgment.

\paragraph{Stage 1: Instruction Following}

This stage focuses on query-dependent criteria. Given a user instruction and its associated context, the LLM judge first derives an acceptance checklist. This checklist encapsulates both explicit constraints (e.g., language, speaker count, topic focus, duration) and implicit expectations inferred from the input. The generated script is then audited against each item, receiving a score of $\{0, 0.5, 1\}$ based on the presence of supporting evidence. These item-level scores are aggregated and normalized to produce a final Instruction Following score (0–100).

\paragraph{Stage 2: Podcast Script Quality}

To capture the nuances of human preference missed by generic checklists, Stage 2 employs a domain-specific rubric designed for \textit{audio-first} content. This 100-point framework assesses three dimensions critical to the podcast medium:

(1) \textbf{Content Substance (45 pts)}: This dimension evaluates the provision of unique insights and perspectives rather than mere factual summarization. It emphasizes the script's ability to engage listeners through deep analysis and emotional resonance, distinguishing high-value synthesis from generic information retrieval.

(2) \textbf{Narrative Engagement (30 pts)}: This dimension assesses narrative flow and audience retention strategies. Unlike the static structure of written essays, it prioritizes auditory engagement techniques—such as opening hooks, effective pacing, and smooth transitions—essential for sustaining listener interest.

(3) \textbf{Conversational Naturalness (25 pts)}: This dimension measures the script's suitability for oral delivery. It rewards conversational phrasing and vivid imagery ("show, don't tell") while penalizing complex syntactic structures or excessive jargon that are difficult to articulate or comprehend aurally.

\paragraph{Score Aggregation}

For each model output, we report the final PodBench score as the average of the Instruction Following score (Stage 1) and the Podcast Script Quality score (Stage 2). We also report dimension-level breakdowns to facilitate detailed analysis.

\section{Experiments}
\subsection{Experimental Setup}
\label{sec:experimental_setup}

\paragraph{Evaluation Protocol.} We use the recommended generation hyper-parameters from the official release websites of baseline LLMs to achieve the best performance when generating responses for PodBench data. For instance, for DeepSeek-R1-0528, the temperate is set to 0.6 and the top-p value is set to 0.95. The maximum new generated tokens is uniformly set to 16,384 tokens. The Claude-4.5-Opus with the `high' effort level is adopted to perform LLM-as-a-judge scoring with the consistent configurations. The temperature is set to 1.0. We apply a top-p value of 0.95 without using top-k sampling. The max new tokens is also set to 16,384 tokens for generating thorough evaluation results. The final input prompt of test samples are established based on instructions and input materials with a fixed prompt template, which is shown in \ref{app:PodcastScriptGenerationPrompt}.

\paragraph{Baseline Models.} We evaluate 34 different LLMs, covering a wide range of types and model sizes. \textit{(1) Proprietary LLMs} including GPT-4o~\cite{openai2024gpt4ocard}, GPT-5.1-instant~\cite{GPT_5_1}, Claude-4.5-Sonnet~\cite{claude_4_5_sonnet} and Gemini-3-pro-preview~\cite{gemini_3_pro}. \textit{(2) Open-source LLMs}. We consider InternLM3-8B-Instruct~\cite{internlm2}, Llama3.1-instruct with 8B and 70B sizes~\cite{grattafiori2024llama3herdmodels}, Llama-4-Scout-17B-16E-Instruct~\cite{llama-4}, Qwen2.5-instruct models of 7B, 32B and 72B sizes~\cite{qwen2025qwen25technicalreport}, Qwen3 models of 1.7B, 4B, 8B, 32B, 30B-A3B and 235B-A22B with hybrid thinking mode~\cite{yang2025qwen3technicalreport}, DeepSeek-V3-0324~\cite{deepseekai2025deepseekv3technicalreport} and DeepSeek-R1-0528 and R1-distill series models~\cite{deepseekai2025deepseekr1incentivizingreasoningcapability}. For Qwen3 models, we follow the official methods to switch between thinking and non-thinking modes. \textit{(3) Writing-enhanced LLMs} including LongWriter~\cite{bai2024longwriter} and LongWriter-Zero~\cite{long_writer_zero}. For R1-style reasoning LLMs, the reasoning content is ignored and we only extract the final podcast scripts from the model responses for evaluation.

\subsection{Results and Analysis}

\paragraph{Main Results.}
Table~\ref{main_results_table} presents the performance of various proprietary and open-source models on PodBench. A persistent gap is observed between meeting instruction constraints and achieving high intrinsic script quality. Several strong systems obtain high Instruction Following (Stage 1) scores yet leave substantial headroom on the Podcast Script Quality rubric (Stage 2). This suggests that satisfying explicit constraints (e.g., language, speaker count, duration) does not automatically translate into engaging, high-quality dialogue.

A second consistent pattern is that the largest deficit lies in Content Substance rather than surface form. Compared to Narrative Engagement and Conversational Naturalness, Content Substance remains the most challenging dimension even for competitive models. This indicates that sustaining grounded, non-trivial analysis throughout a long-form multi-speaker dialogue is significantly harder than maintaining fluency or structure.

We further find that explicit reasoning improves performance, though gains are heterogeneous. Within the Qwen3 family, enabling "thinking mode" boosts both following and quality, with larger relative improvements for mid-sized models (e.g., 14B) than for very large ones (e.g., 235B). This implies that explicit reasoning primarily aids planning and discourse control when model capacity is constrained. Additionally, training methodology can partially substitute for scale: strong open models (e.g., DeepSeek-V3 / DeepSeek-R1-0528) reach performance bands comparable to substantially larger instruct models, highlighting the importance of effective long-context training. Finally, writing-enhanced models (e.g., LongWriter-Zero-32B) show a mismatch: they achieve high Stage 2 quality relative to many open baselines but do not dominate in Stage 1, suggesting that improvements in narrative packaging do not necessarily coincide with stronger constraint satisfaction.

\paragraph{The Influence of Input Length.}

\begin{figure*}[t]
  \centering
  \includegraphics[width=0.8\linewidth]{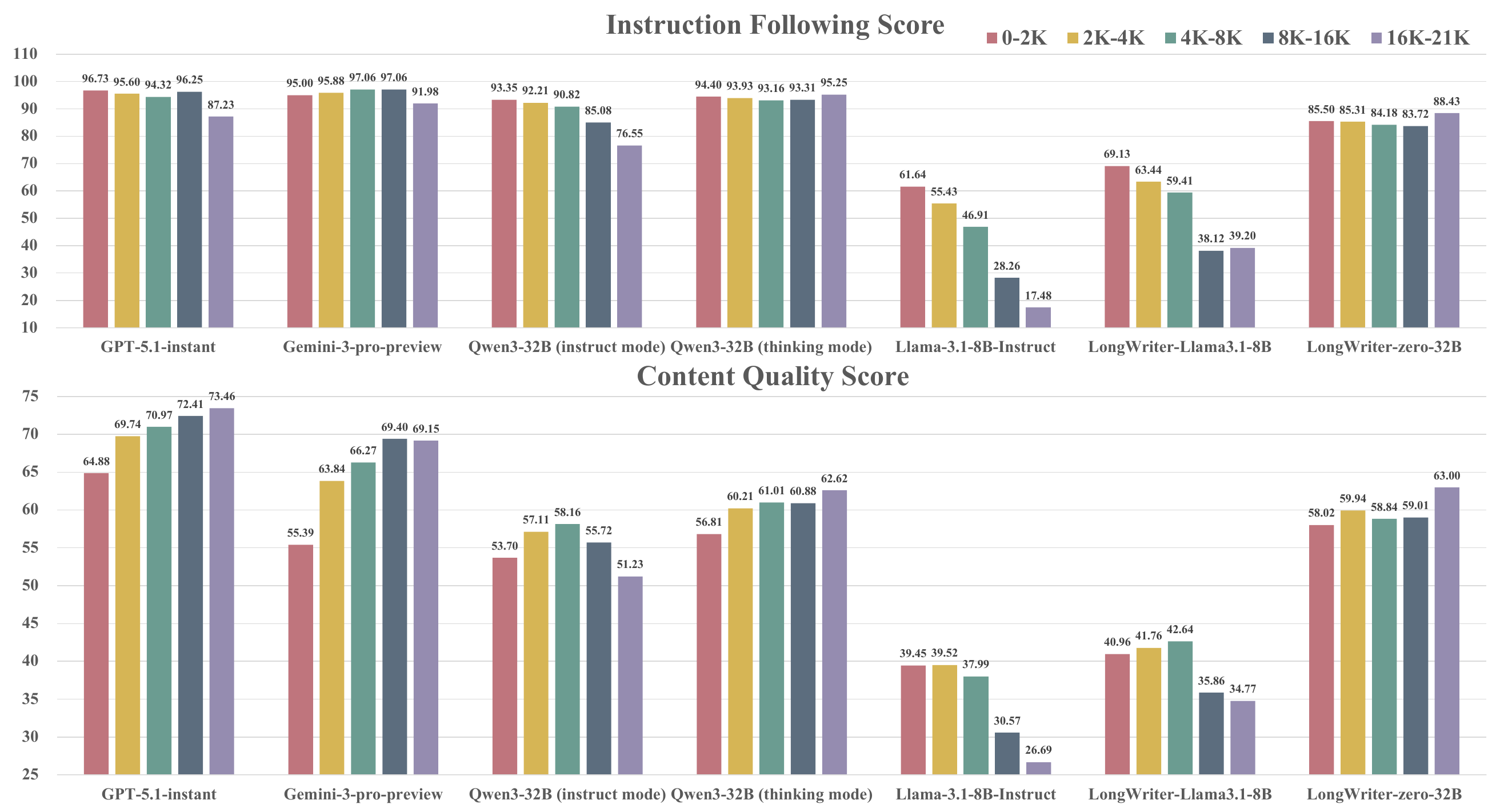}
  \caption{The comparison of model performance on samples of different input length.}
  \label{fig:model_performance_by_input_length}
\end{figure*}

Figure~\ref{fig:model_performance_by_input_length} illustrates how input length affects performance. Analysis reveals that longer inputs pose a more consistent challenge for Instruction Following than for Podcast Script Quality. Most open models exhibit a monotonic decline in adherence as context length approaches 16K--21K tokens, suggesting that extended contexts amplify the difficulty of tracking constraints (e.g., format, topic). In contrast, script quality shows non-monotonic trends: stronger models often leverage richer contexts to enhance quality, whereas weaker models plateau or regress.

Length robustness generally correlates with model capacity, with larger models degrading more gracefully. Notably, explicit reasoning enhances stability across length buckets, particularly for mid-sized models. These gains are more pronounced for adherence than for quality, supporting the hypothesis that explicit planning aids in maintaining structural compliance under long-context settings, while quality remains bounded by the model's intrinsic writing capabilities.

\paragraph{The Influence of Speaker Number.}

\begin{figure}[t]
  \centering
  \includegraphics[width=\linewidth]{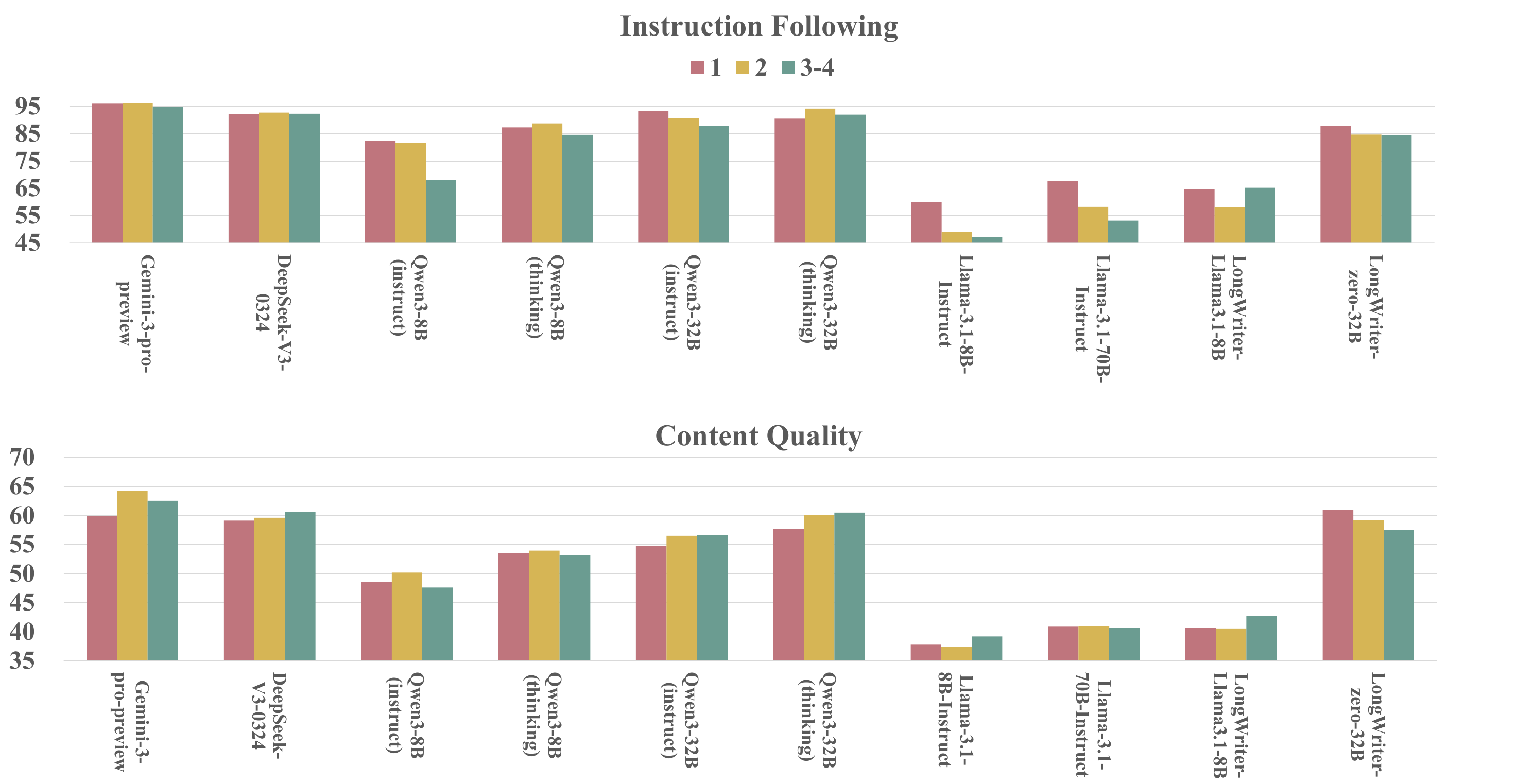}
  \caption{The comparison of model performance on samples of different speaker number. `1', `2' and `3-4' indicates the required speaker number in the user instructions.}
  \label{fig:model_performance_by_speaker_number}
\end{figure}

Figure~\ref{fig:model_performance_by_speaker_number} compares model behavior as the number of speakers increases. We observe a clear \emph{coordination tax} that disproportionately impacts Instruction Following. In 3--4 speaker scenarios, open instruct models exhibit notable instability in role tracking and turn-taking, while script quality remains relatively stable. This suggests that multi-speaker settings primarily strain models’ \emph{constraint management} capacity rather than their core generation fluency.

Proprietary models show greater robustness and often exhibit a slight performance gain with two speakers compared to single-speaker baselines. We hypothesize that dyadic interactions offer structural scaffolding (e.g., Q\&A dynamics) that supports narrative organization without incurring the high coordination overhead of larger speaker sets. Importantly, explicit reasoning mitigates sensitivity to speaker count: models operating in thinking mode remain stable as speaker numbers increase, whereas style-optimized models continue to struggle, indicating that stylistic optimization alone is insufficient for complex coordination.

\paragraph{Validation of Evaluation Metric.} 

To assess the validity of the proposed scoring framework, we analyze the alignment between automated scores and human preference on a held-out set of 200 queries. We compare PodBench against two baseline evaluation protocols to isolate the contributions of domain knowledge and weighting schemes: (1) a Generic Checklist-Based Evaluation, which generates query-specific checklists without domain priors, and (2) a Uniform Domain Rubric, which applies our domain-specific criteria with equal weighting.

As shown in Table~\ref{tab:human_agreement}, progressively enriching the evaluation protocol with domain-specific criteria and weighting leads to improved alignment with human judgments. The results indicate that capturing domain knowledge and relative preference importance contributes to a more reliable approximation of professional human evaluation.

\begin{table}[h]
    \centering
    \small
    \renewcommand{\arraystretch}{1.1}
    \scalebox{0.9}{
    \setlength{\tabcolsep}{4pt}
    \begin{tabular}{lc}
    \toprule
    \textbf{Method} & \textbf{Accuracy (\%)} \\
    \midrule
    \textit{Human Reference} & \\
    Inter-Annotator Agreement (Avg.) & 83.36 \\
    \midrule
    \textit{Automated Evaluators} & \\
    Generic Checklist-Based Evaluation & 79.60 \\
    Uniform Domain Rubric & 83.22 \\
    \textbf{PodBench (Ours)} & \textbf{86.80} \\
    \bottomrule
    \end{tabular}
    }
    \caption{Alignment between human majority preference and different automated evaluation protocols.}
    \label{tab:human_agreement}
\end{table}

\section{Conclusion}

This work introduces PodBench, a comprehensive benchmark for instruction-aware and context-grounded podcast script generation. Our multifaceted evaluation uncovers a critical divergence: while models often achieve high Instruction Following, they struggle to maintain Content Substance, particularly under long-context and multi-speaker conditions. Notably, we find that explicit reasoning significantly enhances robustness against these structural constraints, effectively mitigating the "coordination tax" that degrades standard models. PodBench thus serves as a rigorous testbed to guide future research in long-form, audio-centric conversational AI.

\section*{Limitations}

This work has several limitations.

First, while PodBench reflects common settings in current AI-podcast applications, it does not aim to cover the full diversity of podcast formats or conversational styles. Certain aspects, such as highly improvisational dialogue or audience interaction, are not explicitly modeled.

Second, the evaluation framework relies on LLM-based judging to assess both explicitly checkable constraints and more subjective, audio-oriented quality dimensions, enabling scalable and unified evaluation. While this design allows PodBench to cover a broad range of podcast-relevant criteria, it remains an approximation of human judgment and may not fully capture individual listener preferences.

Third, PodBench focuses on text-level podcast script generation and does not consider downstream audio-related factors, such as speech synthesis quality or prosody. In addition, our experiments evaluate general-purpose language models under standard inference and selected reasoning-enhanced settings, without introducing additional task-specific adaptation or training beyond their original configurations.

\section*{Ethical Considerations}

We discuss the ethical considerations related to the construction and evaluation of PodBench.

\paragraph{Data Risk Control.} To reduce potential risks, we apply both automated and manual filtering procedures during benchmark construction. Input materials and synthesized instructions are screened to remove content involving explicit violence, sensitive or harmful topics, or material inappropriate for general audiences. In addition, we conduct explicit checks to ensure that no personally identifiable information (PII) is included in the benchmark. These checks are performed at multiple stages of data processing, including automatic detection and manual verification by the authors.

\paragraph{Annotator Treatment and Consent.} PodBench does not rely on subjective human annotation for quality labeling. Human involvement is limited to data curation, verification, and consistency checking of automatically constructed samples. These activities are conducted either by trained team members or by professional annotators employed through commercial annotation service providers, all of whom are compensated for their work in accordance with local labor regulations and standard industry practices.

All annotators follow predefined review guidelines and are not required to produce creative content or subjective quality judgments. The annotation process does not involve personally identifying information or sensitive content beyond publicly available or synthetic text. As a result, risks related to annotator consent, compensation, and exposure to harmful material are minimal.

\paragraph{Copyright and Data Licensing.} The input materials used in PodBench are derived from publicly available datasets with explicit usage licenses, as summarized in Table~\ref{tab:dataset_license}. These datasets include both Chinese and English corpora and are released under permissive licenses such as CC BY 4.0 and MIT. PodBench constructs contextual input materials through sampling and processing of licensed sources for evaluation purposes. All data usage adheres to the original licensing terms of the respective datasets.

\paragraph{Potential Risks.}
A potential risk of PodBench lies in the interpretation and use of automated evaluation results. Although we validate strong alignment between PodBench scores and human preferences, the framework should not be treated as a definitive substitute for expert editorial judgment. There is also a risk that benchmark scores may be over-interpreted as indicators of real-world podcast quality beyond the intended research scope. To mitigate these risks, we emphasize that PodBench is designed as a diagnostic research tool, and its results should be considered alongside qualitative analysis and, when appropriate, human evaluation.

\bibliography{custom}

\newpage
\appendix

\section{More Benchmark Information}
\label{sec::more_benchmark_info}

\begin{table}[t]\footnotesize
\centering
\begin{tabular}{c|c|c} \toprule
\textbf{Language} & \textbf{Dataset} & \textbf{License} \\
\midrule
Chinese & \makecell[c]{LongWanjuan \\ \cite{longwanjuan}}& CC BY 4.0 \\
Chinese & \makecell[c]{OpenNewsArchive \\ \cite{OpenDataLab}} & CC BY 4.0 \\
Chinese & \makecell[c]{WanJuan~1.0 \\ \cite{WanJuan_1}} & CC BY 4.0 \\
English & \makecell[c]{WanJuan~CC \\ \cite{WanJuan_CC}} & CC BY 4.0 \\
English & \makecell[c]{Pile-ArXiv \\ \cite{the_pile}} & MIT \\
\bottomrule
\end{tabular}
\caption{Datasets and licenses used for constructing input materials in PodBench.}
\label{tab:dataset_license}
\end{table}

\paragraph{Data Sources and Licenses} Table~\ref{tab:dataset_license} lists the datasets used to construct the contextual input materials in PodBench, together with their corresponding licenses. 
All datasets are publicly available and cover both Chinese and English sources, including long-form documents, news articles, web pages, and scientific papers. 
These sources are used solely to provide contextual grounding for podcast script generation and are not redistributed as part of the benchmark.

\section{More Experimental Results}

\subsection{Detailed Performance by Input Length.} Table~\ref{detailed_performance_by_input_length} provides a granular breakdown of model performance across five input-length buckets, ranging from short contexts (0--2K tokens) to extended contexts (16K--21K tokens). We report scores for both Instruction Following (Stage 1) and Podcast Script Quality (Stage 2), along with the standard deviation (STD) to quantify performance stability across varying lengths. This detailed analysis highlights the impact of context scaling on model robustness, revealing how different architectures and inference modes (e.g., instruct vs. thinking) cope with increasing information load.

\begin{table*}[t]\footnotesize
\centering
\setlength\tabcolsep{2pt}
\renewcommand{\arraystretch}{1.2}
    \scalebox{0.9}{
    \begin{tabular}{l|cccccc|cccccc} \hline
\multirow{2}{*}{\textbf{Method}} & \multicolumn{6}{c|}{\textbf{Instruction Following}} & \multicolumn{6}{c}{\textbf{Content Quality}} \\ \cline{2-13}
 & \multicolumn{1}{c|}{\textbf{0-2K}} & \multicolumn{1}{c|}{\textbf{2K-4K}} & \multicolumn{1}{c|}{\textbf{4K-8K}} & \multicolumn{1}{c|}{\textbf{8K-16K}} & \multicolumn{1}{c|}{\textbf{16K-21K}} & \textbf{STD} & \multicolumn{1}{c|}{\textbf{0-2K}} & \multicolumn{1}{c|}{\textbf{2K-4K}} & \multicolumn{1}{c|}{\textbf{4K-8K}} & \multicolumn{1}{c|}{\textbf{8K-16K}} & \multicolumn{1}{c|}{\textbf{16K-21K}} & \textbf{STD} \\ \hline
\multicolumn{13}{l}{{\cellcolor[rgb]{0.933,0.933,0.933}}\textbf{Proprietary-LLMs}} \\
Claude-4-5-Sonnet & \textbf{96.88} & \textbf{96.66} & 96.40 & 96.47 & \textbf{94.12} & \textbf{1.13} & 58.74 & 63.65 & 64.44 & 64.79 & 70.46 & 4.17 \\
GPT-5.1-instant & 96.73 & 95.60 & 94.32 & 96.25 & 87.23 & 3.91 & \textbf{64.88} & \textbf{69.74} & \textbf{70.97} & \textbf{72.41} & \textbf{73.46} & 3.34 \\
GPT-4o & 90.40 & 89.64 & 89.93 & 87.88 & 83.52 & 2.82 & 53.85 & 57.58 & 57.88 & 57.05 & 55.23 & \textbf{1.72} \\
Gemini-3-pro-preview & 95.00 & 95.88 & \textbf{97.06} & \textbf{97.06} & 91.98 & 2.10 & 55.39 & 63.84 & 66.27 & 69.40 & 69.15 & 5.74 \\
\multicolumn{13}{l}{{\cellcolor[rgb]{0.933,0.933,0.933}}\textbf{Open-source LLMs (instruct mode)}} \\
InternLM3-8B-instruct & 75.92 & 75.72 & 72.86 & 69.03 & 67.72 & \textbf{3.77} & 44.57 & 46.16 & 46.27 & 45.97 & 47.08 & \textbf{0.91} \\
Llama-3.1-8B-Instruct & 61.64 & 55.43 & 46.91 & 28.26 & 17.48 & 18.58 & 39.45 & 39.52 & 37.99 & 30.57 & 26.69 & 5.87 \\
Llama-3.1-70B-Instruct & 70.20 & 62.82 & 57.37 & 39.06 & 36.20 & 14.90 & 41.12 & 42.31 & 41.16 & 37.15 & 35.31 & 3.01 \\
Llama-4-Scout-17B-16E-Instruct & 78.09 & 72.96 & 67.64 & 58.54 & 55.50 & 9.51 & 43.86 & 45.60 & 44.57 & 42.51 & 41.54 & 1.62 \\
Qwen2.5-7B-Instruct & 76.09 & 71.44 & 66.90 & 60.10 & 48.05 & 10.93 & 42.47 & 44.79 & 43.74 & 43.94 & 39.46 & 2.08 \\
Qwen2.5-32B-Instruct & 82.45 & 79.78 & 75.79 & 71.19 & 66.80 & 6.33 & 46.05 & 47.44 & 47.17 & 45.84 & 45.38 & 0.89 \\
Qwen2.5-72B-Instruct & 85.95 & 84.11 & 83.65 & 78.85 & 74.20 & 4.78 & 46.65 & 49.70 & 49.86 & 48.95 & 48.62 & 1.28 \\
Qwen3-1.7B & 63.91 & 58.16 & 54.96 & 48.67 & 32.95 & 11.86 & 40.89 & 41.68 & 40.48 & 38.13 & 35.08 & 2.68 \\
Qwen3-4B & 83.60 & 79.19 & 73.05 & 67.87 & 69.77 & 6.58 & 46.77 & 48.69 & 48.09 & 45.95 & 42.92 & 2.26 \\
Qwen3-8B & 86.25 & 84.42 & 78.78 & 68.92 & 67.28 & 8.71 & 48.47 & 51.67 & 50.72 & 46.26 & 45.31 & 2.75 \\
Qwen3-14B & 90.85 & 88.38 & 86.49 & 81.74 & 70.95 & 7.86 & 49.53 & 53.30 & 53.55 & 51.12 & 46.62 & 2.87 \\
Qwen3-32B & \textbf{93.35} & 92.21 & 90.82 & 85.08 & \textbf{76.55} & 6.95 & 53.70 & 57.11 & 58.16 & 55.72 & \textbf{51.23} & 2.77 \\
Qwen3-30B-A3B & 87.65 & 88.08 & 85.00 & 83.76 & 76.00 & 4.87 & 49.08 & 52.70 & 53.41 & 53.43 & 49.77 & 2.09 \\
DeepSeek-V3-0324 & 92.81 & \textbf{93.55} & \textbf{93.77} & \textbf{92.51} & 55.42 & 16.89 & \textbf{54.58} & \textbf{60.29} & \textbf{61.99} & \textbf{61.39} & 46.58 & 6.51 \\
\multicolumn{13}{l}{{\cellcolor[rgb]{0.933,0.933,0.933}}\textbf{Open-source LLMs (thinking mode)}} \\
Qwen3-1.7B & 68.20 & 63.83 & 61.14 & 56.05 & 56.06 & 5.21 & 42.83 & 42.92 & 42.10 & 40.50 & 39.92 & 1.37 \\
Qwen3-4B & 86.27 & 84.70 & 81.43 & 78.23 & 70.61 & 6.21 & 49.14 & 52.03 & 51.97 & 50.31 & 47.00 & 2.11 \\
Qwen3-8B & 89.13 & 89.48 & 86.71 & 87.38 & 84.22 & 2.12 & 51.73 & 54.68 & 54.14 & 53.75 & 53.46 & 1.12 \\
Qwen3-14B & 92.42 & 91.90 & 91.88 & 90.98 & 89.34 & 1.21 & 53.55 & 57.46 & 58.83 & 56.62 & 54.85 & 2.09 \\
Qwen3-32B & \textbf{94.40} & 93.93 & 93.16 & 93.31 & \textbf{95.25} & \textbf{0.85} & \textbf{56.81} & 60.21 & 61.01 & 60.88 & \textbf{62.62} & 2.15 \\
Qwen3-30B-A3B & 90.63 & 89.51 & 90.49 & 89.60 & 87.83 & 1.12 & 51.05 & 55.34 & 55.64 & 55.83 & 58.31 & 2.62 \\
DeepSeek-R1-Distill-Llama-8B & 72.23 & 69.30 & 65.34 & 58.00 & 55.68 & 7.12 & 43.37 & 45.99 & 45.78 & 44.45 & 45.31 & \textbf{1.08} \\
DeepSeek-R1-Distill-Qwen-7B & 54.18 & 48.42 & 44.38 & 15.14 & 10.69 & 20.13 & 36.62 & 38.12 & 37.85 & 29.69 & 24.36 & 6.08 \\
DeepSeek-R1-Distill-Qwen-32B & 83.24 & 81.15 & 77.74 & 76.52 & 82.85 & 3.03 & 45.46 & 48.61 & 48.22 & 48.96 & 48.77 & 1.45 \\
DeepSeek-R1-0528-Qwen3-8B & 87.72 & 86.20 & 86.71 & 81.29 & 81.55 & 3.04 & 48.50 & 51.82 & 52.24 & 51.74 & 51.15 & 1.50 \\
DeepSeek-R1-0528 & 93.56 & \textbf{94.78} & \textbf{95.40} & \textbf{96.68} & 46.32 & 21.85 & 56.32 & \textbf{61.10} & \textbf{63.47} & \textbf{64.65} & 50.17 & 5.94 \\
\multicolumn{13}{l}{{\cellcolor[rgb]{0.933,0.933,0.933}}\textbf{Writing-enhanced LLMs}} \\
LongWriter-Llama3.1-8B & 69.13 & 63.44 & 59.41 & 38.12 & 39.20 & 14.30 & 40.96 & 41.76 & 42.64 & 35.86 & 34.77 & 3.61 \\
LongWriter-GLM4-9B & 73.99 & 66.90 & 64.77 & 57.39 & 58.77 & 6.69 & 42.83 & 44.71 & 43.90 & 43.81 & 42.23 & \textbf{0.97} \\
LongWriter-zero-32B & \textbf{85.50} & \textbf{85.31} & \textbf{84.18} & \textbf{83.72} & \textbf{88.43} & \textbf{1.84} & \textbf{58.02} & \textbf{59.94} & \textbf{58.84} & \textbf{59.01} & \textbf{63.00} & 1.93 \\ \hline
\end{tabular}
    }
    \caption{Detailed model performance on samples of different input length. `STD' represents the standard deviation.  The best results in each model category are highlighted in bold.}
    \label{detailed_performance_by_input_length}
\end{table*}

\subsection{Detailed Performance by Speaker Number.} Table~\ref{detailed_performance_by_speaker_num} presents a comparative analysis of model performance across varying speaker counts (1, 2, and 3--4 speakers). We report scores for Instruction Following and Podcast Script Quality, alongside the standard deviation (STD) to measure stability. This breakdown highlights the "coordination tax" associated with multi-speaker generation, revealing how different models manage the increased structural complexity of role allocation and turn-taking while striving to maintain content quality.

\begin{table*}[t]\footnotesize
\centering
\setlength\tabcolsep{8pt}
\renewcommand{\arraystretch}{1.2}
    \scalebox{1.0}{
    \begin{tabular}{l|cccc|cccc} \hline
\multirow{2}{*}{\textbf{Method}} & \multicolumn{4}{c|}{\textbf{Instruction Following}} & \multicolumn{4}{c}{\textbf{Content Quality}} \\ \cline{2-9}
 & \multicolumn{1}{c|}{\textbf{1}} & \multicolumn{1}{c|}{\textbf{2}} & \multicolumn{1}{c|}{\textbf{3-4}} & \textbf{STD} & \multicolumn{1}{c|}{\textbf{1}} & \multicolumn{1}{c|}{\textbf{2}} & \multicolumn{1}{c|}{\textbf{3-4}} & \textbf{STD} \\ \hline
\multicolumn{9}{l}{{\cellcolor[rgb]{0.945,0.945,0.945}}\textit{Proprietary-LLMs}} \\
Claude-4-5-Sonnet & 94.52 & \textbf{96.85} & \textbf{95.16} & 1.20 & 60.85 & 63.51 & 62.51 & 1.34 \\
GPT-5.1-instant & 94.60 & 95.66 & 94.74 & \textbf{0.58} & \textbf{68.07} & \textbf{69.78} & \textbf{69.68} & \textbf{0.96} \\
GPT-4o & 88.38 & 89.70 & 86.89 & 1.41 & 54.23 & 57.09 & 56.85 & 1.59 \\
Gemini-3-pro-preview & \textbf{96.04} & 96.18 & 94.79 & 0.77 & 59.87 & 64.29 & 62.55 & 2.23 \\
\multicolumn{9}{l}{{\cellcolor[rgb]{0.945,0.945,0.945}}\textit{Open-source LLMs (instruct mode)}} \\
InternLM3-8B-instruct & 72.68 & 74.50 & 67.54 & 3.61 & 43.97 & 46.12 & 44.87 & 1.08 \\
Llama-3.1-8B-Instruct & 59.88 & 49.01 & 47.04 & 6.92 & 37.77 & 37.38 & 39.20 & 0.96 \\
Llama-3.1-70B-Instruct & 67.73 & 58.21 & 53.18 & 7.39 & 40.87 & 40.93 & 40.66 & \textbf{0.14} \\
Llama-4-Scout-17B-16E-Instruct & 72.30 & 70.19 & 66.64 & 2.86 & 43.39 & 44.66 & 43.89 & 0.64 \\
Qwen2.5-7B-Instruct & 72.56 & 69.24 & 62.72 & 5.01 & 42.10 & 44.24 & 42.13 & 1.23 \\
Qwen2.5-32B-Instruct & 80.50 & 78.11 & 70.43 & 5.26 & 45.28 & 46.97 & 47.11 & 1.02 \\
Qwen2.5-72B-Instruct & 83.46 & 83.57 & 79.36 & 2.40 & 47.00 & 49.29 & 48.26 & 1.15 \\
Qwen3-1.7B & 65.43 & 55.96 & 53.13 & 6.44 & 42.84 & 40.46 & 39.80 & 1.60 \\
Qwen3-4B & 79.65 & 77.21 & 65.84 & 7.37 & 46.30 & 47.85 & 47.09 & 0.78 \\
Qwen3-8B & 82.46 & 81.51 & 68.06 & 8.05 & 48.56 & 50.20 & 47.62 & 1.31 \\
Qwen3-14B & 89.45 & 87.24 & 81.47 & 4.12 & 50.03 & 52.38 & 52.77 & 1.48 \\
Qwen3-32B & \textbf{93.40} & 90.65 & 87.77 & 2.82 & 54.82 & 56.52 & 56.60 & 1.01 \\
Qwen3-30B-A3B & 88.60 & 86.87 & 78.30 & 5.52 & 51.41 & 52.39 & 51.66 & 0.51 \\
DeepSeek-V3-0324 & 92.18 & \textbf{92.78} & \textbf{92.32} & \textbf{0.31} & \textbf{59.13} & \textbf{59.59} & \textbf{60.57} & 0.74 \\
\multicolumn{9}{l}{{\cellcolor[rgb]{0.945,0.945,0.945}}\textit{Open-source LLMs (thinking mode)}} \\
Qwen3-1.7B & 69.85 & 62.54 & 54.86 & 7.50 & 44.34 & 42.10 & 42.49 & 1.20 \\
Qwen3-4B & 84.87 & 83.27 & 77.36 & 3.96 & 50.25 & 51.26 & 50.70 & 0.51 \\
Qwen3-8B & 87.40 & 88.80 & 84.60 & 2.14 & 53.56 & 53.97 & 53.15 & 0.41 \\
Qwen3-14B & 90.24 & 92.12 & 88.93 & \textbf{1.60} & 55.48 & 56.95 & 57.26 & 0.95 \\
Qwen3-32B & 90.57 & 94.19 & \textbf{91.96} & 1.83 & 57.66 & 60.08 & 60.51 & 1.54 \\
Qwen3-30B-A3B & 90.18 & 90.22 & 84.46 & 3.31 & 54.21 & 54.89 & 53.98 & 0.47 \\
DeepSeek-R1-Distill-Llama-8B & 72.73 & 66.76 & 61.92 & 5.41 & 45.49 & 45.27 & 44.13 & 0.73 \\
DeepSeek-R1-Distill-Qwen-7B & 52.08 & 42.12 & 36.82 & 7.75 & 36.74 & 36.16 & 36.16 & \textbf{0.33} \\
DeepSeek-R1-Distill-Qwen-32B & 78.53 & 80.72 & 73.37 & 3.77 & 45.68 & 48.29 & 47.62 & 1.36 \\
DeepSeek-R1-0528-Qwen3-8B & 87.11 & 86.12 & 77.23 & 5.44 & 47.67 & 51.67 & 50.62 & 2.07 \\
DeepSeek-R1-0528 & \textbf{92.35} & \textbf{94.66} & 91.00 & 1.85 & \textbf{59.90} & \textbf{61.18} & \textbf{62.40} & 1.25 \\
\multicolumn{9}{l}{{\cellcolor[rgb]{0.945,0.945,0.945}}\textit{Writing-enhanced LLMs}} \\
LongWriter-Llama3.1-8B & 64.59 & 58.13 & 65.20 & 3.92 & 40.66 & 40.56 & 42.68 & 1.20 \\
LongWriter-GLM4-9B & 72.22 & 65.83 & 60.36 & 5.94 & 43.41 & 44.22 & 42.30 & 0.96 \\
LongWriter-zero-32B & \textbf{88.00} & \textbf{84.66} & \textbf{84.55} & \textbf{1.96} & \textbf{61.02} & \textbf{59.25} & \textbf{57.53} & \textbf{1.75} \\ \hline
\end{tabular}
    }
    \caption{Detailed model performance on samples of different speaker number. `1', `2' and `3-4' indicates the required speaker number in the user instructions. `STD' represents the standard deviation.  The best results in each model category are highlighted in bold.}
    \label{detailed_performance_by_speaker_num}
\end{table*}

\section{Prompts Used In PodBench}
\subsection{Podcast Script Generation Prompt}
\label{app:PodcastScriptGenerationPrompt}

The prompt below is used to generate podcast scripts from user instructions and contextual input materials, as described in Section~\ref{sec:experimental_setup}. It enforces strict source grounding, instruction following, and audio-oriented dialogue design.

\begin{figure*}[t]
\centering
\begin{adjustbox}{max totalsize={\textwidth}{0.95\textheight},center}

\begin{tcolorbox}[colback=gray!5,
    colframe=gray!80,
    width=\textwidth,
    arc=2mm, auto outer arc,
    title={\textbf{Podcast Script Generation Prompt}},
    breakable, enhanced jigsaw,
    before upper={\parindent15pt\noindent}]
You are a top-tier podcast script creation expert, specializing in transforming raw source materials into vivid, insightful, and engaging podcast dialogue scripts. You are skilled at deconstructing complex information, designing natural conversational rhythms, and balancing intellectual depth with listener engagement.

\textbf{Task.} Your task is to transform the provided \texttt{<Input Materials>} into a high-quality podcast script that strictly follows the \textbf{Podcast Script Requirements} below, while prioritizing the \texttt{<User Requirements>}.

\textbf{Core Requirements.}
\begin{itemize}
    \item \textbf{User requirements take priority.} If conflicts arise, always defer to user-specified requirements.
    \item \textbf{Source grounding.} All facts, viewpoints, and data must originate exclusively from the input materials. External knowledge or fabrication is strictly prohibited.
    \item \textbf{Role configuration.} Generate dialogue according to the specified speaker roles and interaction mode, reflecting distinct speaking styles without explicitly naming roles in the dialogue.
    \item \textbf{Audio-first design.} Ensure the script is natural to listen to, with smooth transitions, conversational fillers when appropriate, and coherent dialogue flow.
    \item \textbf{Depth and accessibility.} Deliver informative and insightful content while remaining accessible and engaging for a general audience.
    \item \textbf{Multi-speaker interaction.} For multi-host podcasts, encourage progressive discussion rather than simple question–answer patterns.
    \item \textbf{Faithful integration.} Do not explicitly cite source documents (e.g., ``according to the article''). Internalize information as the speakers’ own discussion.
    \item \textbf{Length control.} Approximately 300 characters correspond to one minute of podcast content.
\end{itemize}

\textbf{User Requirements:}
\begin{verbatim}
{user_intent}
\end{verbatim}

\textbf{Script Role Configuration:}
\begin{verbatim}
{role_description}
\end{verbatim}

\textbf{Input Materials:}
\begin{verbatim}
{content}
\end{verbatim}

\textbf{Output Format (example with two hosts):}
\begin{verbatim}
Host 1:
Host 2:
Host 1:
Host 2:
...
\end{verbatim}

The output must strictly follow the specified format and must not include any explanations or metadata.
\end{tcolorbox}
\end{adjustbox}
\end{figure*}

\subsection{Stage-1 Evaluation Prompt (Instruction Following)} 
\label{app:Stage1EvaluationPrompt}

The prompt presented below facilitates the Stage 1 assessment of \textit{Instruction Following}, as detailed in Section~\ref{sec:evaluation_framework}. Unlike generic quality evaluations, this prompt focuses strictly on constraint satisfaction. It directs the LLM judge to (i) derive a dynamic acceptance checklist capturing both explicit requirements and implicit expectations from the input context, and (ii) conduct a rigorous, evidence-based audit of the script against each item. The protocol includes specific mechanisms for validating duration constraints via length conversion and assigning graded satisfaction scores.

\begin{figure*}[t]
\centering
\begin{adjustbox}{max totalsize={\textwidth}{0.95\textheight},center}
\begin{tcolorbox}[colback=gray!5,
    colframe=gray!80,
    width=\textwidth,
    arc=2mm, auto outer arc,
    title={\textbf{Stage-1 Evaluation Prompt (Instruction Acceptance)}},
    breakable, enhanced jigsaw,
    before upper={\parindent15pt\noindent}]
\textbf{Role.} You are a \textbf{User Intent Acceptance Expert}. Your job is not to ``read'' the script, but to \textbf{acceptance-test} it. You must strictly follow a two-step mindset:

\begin{enumerate}
    \item \textbf{Legislator mindset:} derive a \textbf{[Acceptance Checklist]} from the user’s full conversation context, explicit instructions, and salient implicit intent.
    \item \textbf{Judge mindset:} use this checklist to conduct a strict, item-by-item audit of the \textbf{generated podcast script}.
\end{enumerate}

\textbf{Goal.} Based on the complete user context, construct acceptance criteria that cover both \textbf{explicit requirements} (e.g., duration, topic) and \textbf{notable implicit requirements} (e.g., information density, style), and use these criteria as the \emph{only} basis to verify whether the generated \textbf{podcast script} meets the user’s intent.

\textbf{Evaluation Rules.}
\begin{itemize}
    \item \textbf{Itemized judgments.} You must make an independent decision for \emph{each} inferred criterion; do not provide a generic summary.
    \item \textbf{Evidence-based.} Avoid subjective speculation. Every judgment must be supported by direct textual evidence found in the script.
\end{itemize}

\textbf{Duration \& Length Conversion.}
\begin{itemize}
    \item \textbf{Conversion.} $\sim$1 minute of podcast $\approx$ 300 Chinese characters (e.g., ``5 minutes'' $\Rightarrow$ $\sim$1500 characters).
    \item \textbf{Tolerance.} If the user does not impose a strict hard limit, allow \textbf{$\pm5\%$} as a reasonable margin.
\end{itemize}

\textbf{Inputs.}

\textbf{(1) User Instructions:}
\begin{verbatim}
{queries}
\end{verbatim}

\textbf{(2) Podcast Script to be Audited:}
\begin{verbatim}
{podcast_script}
\end{verbatim}

\textbf{Audit Logic (must be executed in this order).}

\textbf{Step 1: Build the Acceptance Checklist.}  
You must first generate an ``acceptance standard'' by analyzing the user instructions, including:

\begin{enumerate}
    \item \textbf{Explicit instructions:} requirements directly stated by the user (e.g., ``10 minutes'', ``be humorous'', ``cover topic X'').
    \item \textbf{Implicit intent:} requirements not stated verbatim but strongly implied by context. First identify the podcast genre (e.g., tech news, lifestyle, paper walkthrough, entertainment gossip), then infer the likely expectations.
    \begin{itemize}
        \item \textit{Length/depth fit:} If the user \textbf{does not specify duration}, infer an appropriate depth based on the \textbf{amount of input materials}.
        \begin{itemize}
            \item \textit{Many materials + no hard limit} $\rightarrow$ the script should be detailed and depth-appropriate (not a short summary of a few hundred characters).
            \item \textit{Few materials + no hard limit} $\rightarrow$ allow moderate expansion, but avoid being overly short.
        \end{itemize}
        \item \textit{Audience fit:} infer the expected information density and level of technicality based on the topic (e.g., medical vs.\ entertainment).
    \end{itemize}
\end{enumerate}

\textbf{Step 2: Execute Acceptance Testing.}  
Using the checklist from Step 1, locate evidence in the script and audit each item.

\begin{itemize}
    \item \textbf{Strict duration audit.} Apply the ``1 minute $\approx$ 300 characters'' rule and the tolerance policy above.
    \item \textbf{Evidence-based enforcement.} If the user requests ``focus on A'' but the script only mentions A without meaningful development, mark it as \textbf{Not satisfied (0)}.
\end{itemize}

\textbf{Scoring.}
\begin{itemize}
    \item \textbf{1 (Fully satisfied):} the script perfectly meets the checklist item.
    \item \textbf{0.5 (Partially satisfied):} the script attempts the item but executes it inadequately (e.g., touches key points without depth).
    \item \textbf{0 (Not satisfied):} the script ignores or violates the item.
\end{itemize}

\textbf{Output Requirements.}
\begin{itemize}
    \item Output \textbf{only one JSON object}. \textbf{No Markdown} and no extra text.
    \item Follow the schema below. \texttt{instruction\_point} must be a \textbf{specific inferred requirement}. Use prefixes such as \texttt{[Explicit/Original]}, \texttt{[Explicit/Refined]}, or \texttt{[Implicit/Inferred]}.
\end{itemize}

\textbf{JSON Output Schema:}

\emph{(Omitted for brevity; the JSON schema remains unchanged.)}

\end{tcolorbox}
\end{adjustbox}
\end{figure*}

\subsection{Stage-2 Evaluation Prompt (Podcast Script Quality)} \label{app:Stage2EvaluationPrompt}

The prompt presented below facilitates the Stage 2 assessment of \textit{Podcast Script Quality}, as detailed in Section~\ref{sec:evaluation_framework}. Adopting an \emph{audio-first} perspective, it requires the LLM judge to provide \emph{evidence-based} critiques supported by verbatim citations from the generated script. The evaluation applies a 100-point rubric across three core dimensions: Content Substance, Narrative Engagement, and Conversational Naturalness. The evaluator is instructed to identify the script type (e.g., monologue, dialogue, narrative feature) prior to scoring and must output the final evaluation as a valid JSON object.

\clearpage
\begin{figure*}[t]
\centering
\begin{adjustbox}{max totalsize={\textwidth}{0.95\textheight},center}
\begin{tcolorbox}[colback=gray!5,
    colframe=gray!80,
    width=14cm,
    arc=2mm, auto outer arc,
    title={\textbf{Stage-2 Evaluation Prompt (Podcast Quality Scoring)}},
    breakable, enhanced jigsaw,
    before upper={\parindent15pt\noindent}]
\textbf{Role.} You are a senior podcast content reviewer with 10+ years of experience in podcast production and supervision. You are deeply familiar with theories in communication, narrative studies, and cognitive psychology. You have reviewed 500+ podcast shows, and you are known for an incisive style: \emph{``trust what you hear, keep logic rigorous, and cut to the core pain points.''}

\textbf{Goal.} Using the \textbf{systematic evaluation framework} below, provide a professional, objective, and multi-dimensional quality assessment of the given podcast script.

\textbf{Core Principles (must be strictly followed).}
\begin{enumerate}
    \item \textbf{Audio-first.} Always simulate the listener’s perspective; all judgments must be based on \emph{listenability} rather than \emph{readability}.
    \item \textbf{Evidence-based.} Every comment must quote the script verbatim as evidence (wrap quoted text with [...]). Empty or generic critiques are forbidden.
    \item \textbf{No ``nice-guy'' scoring.} Be strict. If the script is mediocre, say so plainly. A perfect score (100) is reserved only for industry textbook-level work.
    \item \textbf{Type adaptation.} Before scoring, identify the script type (e.g., knowledge monologue, two-person talk, narrative feature), and adjust scoring emphasis accordingly.
    \item \textbf{Format compliance.} Your final output must be valid JSON.
\end{enumerate}

\textbf{Evaluation Framework (Total: 100 points).}

\textbf{I. Depth \& Value (45 points)}

\textbf{1. Content value \& depth of insight (12 points)}\\
\textbf{Scoring guidelines:}
\begin{itemize}
    \item \textbf{10--12:} Delivers an ``aha'' cognitive leap. Makes complex ideas accessible, or extracts deep mechanisms / unique perspectives from familiar phenomena. Strong urge to share after listening. \textit{Anchor example: Explains ``sunk cost'' via a ``buffet mindset'' analogy—instant understanding and shareable.}
    \item \textbf{7--9:} Reaches ``why'' and ``how''; provides actionable methods or logically coherent explanations. Offers some new knowledge, but with less surprise. \textit{Anchor example: Summarizes ``5 tips for efficient meetings''—useful but not novel.}
    \item \textbf{4--6:} Mostly integrates public information; touches surface reasons with shallow analysis. Correct but unoriginal—``true but empty.'' \textit{Anchor example: Says ``sleep matters'' but only repeats ``get 8 hours.''}
    \item \textbf{0--3:} Pure ``what it is'' listing; lacks a core viewpoint, or is outdated / highly homogeneous. \textit{Anchor example: Wikipedia-style biography narration.}
\end{itemize}

\textbf{2. Argument quality \& credibility (10 points)}\\
\textbf{Scoring guidelines:}
\begin{itemize}
    \item \textbf{9--10:} Clear claims with diverse evidence (data / cases / studies / experts). Logic chain is tight. Proactively presents counterarguments or admits limitations (critical thinking).
    \item \textbf{6--8:} Has claims and evidence; mostly self-consistent. But evidence is single-type (e.g., only personal anecdotes) or depth is moderate.
    \item \textbf{3--5:} Vague claims; weak linkage between evidence and claims; logical jumps or circular reasoning.
    \item \textbf{0--2:} Claims-first without evidence; serious logical fallacies (e.g., reversed causality, slippery slope, survivorship bias).
\end{itemize}

\textbf{3. Information richness \& signal-to-noise ratio (8 points)}\\
\textbf{Scoring guidelines:}
\begin{itemize}
    \item \textbf{7--8:} Appropriate density (3--4 effective information points per 200 Chinese characters). No fluff. Facts/data are accurate and source-marked. Terms are used properly and explained.
    \item \textbf{5--6:} Adequate information, with minor redundant buildup. Core facts are accurate, but some data lack support.
    \item \textbf{3--4:} Too much fluff; repetitive phrasing; low effective information density.
    \item \textbf{0--2:} Information overload; jargon stacking causes listening fatigue.
\end{itemize}

\textbf{4. Perspective diversity \& complexity (8 points)}\\
\textbf{Scoring guidelines:}
\begin{itemize}
    \item \textbf{7--8:} Presents contradictions, gray areas, and trade-offs; integrates multi-disciplinary perspectives (history / psychology / economics, etc.); avoids black-and-white framing.
    \item \textbf{5--6:} Attempts multiple perspectives but with average integration; or leans to one stance while staying reasonably objective.
    \item \textbf{3--4:} Single perspective; little reflection; prone to bias.
    \item \textbf{0--2:} Over-simplified, emotion-driven output; lacks rational analysis.
\end{itemize}

\textbf{5. Emotional resonance \& human warmth (7 points)}\\
\textbf{Scoring guidelines:}
\begin{itemize}
    \item \textbf{6--7:} Uses concrete stories to carry abstract ideas; evokes specific emotions (wonder / warmth / reflection); touches universal values or human dilemmas.
    \item \textbf{4--5:} Has emotional expression and some resonance, but feels formulaic.
    \item \textbf{2--3:} Dry and mechanical; like reading a manual.
    \item \textbf{0--1:} Preachy or forced melodrama; hollow sentimentality.
\end{itemize}

\textbf{II. Structure \& Narrative Design (30 points)}

\textbf{6. Opening hook (6 points)}\\
\textbf{Scoring guidelines:}
\begin{itemize}
    \item \textbf{5--6:} Immediately grabs attention via suspense, conflict, surprising facts, counterintuitive questions, or vivid scenes. Clearly promises listener value (knowledge/experience). \textit{Anchor: ``Did you know 90\% of the time you waste each day is enough to learn a new language?''}
    \item \textbf{3--4:} Conventional but relevant intro; listenable but lacks urgency; unclear value promise. \textit{Anchor: ``Today we’ll talk about time management...''}
    \item \textbf{0--2:} Long self-intros, greetings, background padding; flat and aimless. \textit{Anchor: ``Hello, welcome to X, I’m Y, the weather is nice...''}
\end{itemize}

\textbf{7. Overall structure \& logical flow (10 points)}\\
\textbf{Scoring guidelines:}
\begin{itemize}
    \item \textbf{9--10:} Clear structure pattern (e.g., SCQA, hero’s journey, timeline) with golden proportion (opening 5--10\%, body 75--85\%, ending 10--15\%). Smooth transitions and clear ``signposts''.
    \item \textbf{6--8:} Complete structure and mostly reasonable pacing; transitions are mostly natural with occasional stiffness.
    \item \textbf{3--5:} Loose structure; weak linkage; imbalanced proportions; lacks signposts and listeners may get lost.
    \item \textbf{0--2:} Chaotic; no main thread; highly fragmented.
\end{itemize}

\textbf{Common structure templates (reference).}\\
SCQA: Situation $\rightarrow$ Complication $\rightarrow$ Question $\rightarrow$ Answer.\\
Pyramid: Core claim $\rightarrow$ sub-claims 1/2/3 $\rightarrow$ summary.\\
Timeline: Start $\rightarrow$ development $\rightarrow$ turning point $\rightarrow$ climax $\rightarrow$ ending.\\
Contrast: Phenomenon A vs.\ B $\rightarrow$ deeper causes $\rightarrow$ conclusion.

\textbf{8. Narrative rhythm \& tension design (8 points)}\\
\textbf{Scoring guidelines:}
\begin{itemize}
    \item \textbf{7--8:} Alternates dense-info segments, story segments, and reflective pauses; has an emotional curve; uses suspense/foreshadowing to maintain anticipation.
    \item \textbf{5--6:} Basic rhythm; not too dull, but lacks clear peaks or tension design.
    \item \textbf{3--4:} Linear, flat rhythm; easy to lose attention.
    \item \textbf{0--2:} Extremely dragging with excessive fluff, or too fast causing cognitive overload.
\end{itemize}

\textbf{9. Ending completeness \& elevation (6 points)}\\
\textbf{Scoring guidelines:}
\begin{itemize}
    \item \textbf{5--6:} Concise callback to the core idea (one-sentence takeaway); elevates to values/insights; offers a clear CTA or a question; has a memorable closing line.
    \item \textbf{3--4:} Has summary and closure but is conventional and less memorable.
    \item \textbf{1--2:} Rushed ending; weak summary; no closure.
    \item \textbf{0:} Abrupt stop; incoherent.
\end{itemize}

\textbf{III. Language \& Communication Effect (25 points)}

\textbf{10. Colloquialness \& listenability (15 points)}\\
\textbf{Scoring guidelines:}
\begin{itemize}
    \item \textbf{13--15:} Fully conversational; short, punchy sentences; natural spoken connectors; immediate explanations for terms with everyday analogies; strategic repetition of key information. Average sentence length $\le$ 20 Chinese characters, and avoids formal written connectors such as ``although... but...'' and ``because... therefore...''.
    \item \textbf{9--12:} Mostly conversational with occasional written phrasing; some sentences are long and require breath control.
    \item \textbf{5--8:} Many long/complex sentences and formal connectors; awkward to read aloud and hard to follow by listening.
    \item \textbf{0--4:} Pure written-article style; jargon stacking; deeply nested logic; extremely high listening cognitive load.
\end{itemize}

\textbf{11. Vividness \& imagery (10 points)}\\
\textbf{Scoring guidelines:}
\begin{itemize}
    \item \textbf{9--10:} Strong ``show, don’t tell'' writing; uses multi-sensory details; precise verbs/nouns; excellent metaphors/analogies that concretize abstractions.
    \item \textbf{6--8:} Has concrete examples/details; metaphors are appropriate and aid comprehension.
    \item \textbf{3--5:} Relies on adjectives and abstract nouns (e.g., ``very difficult'', ``for various reasons''); lacks supporting details.
    \item \textbf{0--2:} Dull concept stacking with no imagery.
\end{itemize}

\textbf{Output Format (JSON only).}  
You must output strictly in the following JSON structure. Do not output anything outside the JSON.

\emph{(Omitted for brevity; the JSON schema remains unchanged.)}

\textbf{Podcast Script to Evaluate:}
\begin{verbatim}
{podcast_script}
\end{verbatim}
\end{tcolorbox}
\end{adjustbox}
\end{figure*}
\clearpage

\end{document}